\documentclass{article}

    \usepackage[preprint, nonatbib]{neurips_data_2024}

\usepackage[utf8]{inputenc} 
\usepackage[T1]{fontenc}    
\usepackage{hyperref}       
\usepackage{url}            
\usepackage{booktabs}       
\usepackage{amsfonts}       
\usepackage{nicefrac}       
\usepackage{microtype}      
\usepackage{xcolor}         

\usepackage{geometry}
\usepackage{graphicx}

\usepackage{csquotes}
\usepackage[
   style=authoryear-ibid, 
   backend=biber,
   style=apa,
   maxcitenames=2,
   uniquelist=false,
]{biblatex}
\usepackage{hyperref}
\hypersetup{
    colorlinks=true,
    linkcolor=magenta,
    filecolor=magenta,      
    urlcolor=magenta,
    pdftitle={MOCHI},
    pdfpagemode=FullScreen,
    }

\providecommand\yutong[1]{\textcolor{purple}{[Yutong: #1]}}
\providecommand\stephanie[1]{\textcolor{teal}{[Stephanie: #1]}}

\title{Evaluating Multiview Object Consistency \\in Humans and Image Models}

\author{%
  Tyler Bonnen$^1$ 
  \hspace{5mm}
  Stephanie Fu$^1$ 
  \hspace{5mm}
  Yutong Bai$^1$
  \hspace{5mm}
  Thomas O'Connell$^2$ \\
  \textbf{Yoni Friedman$^2$}
  \hspace{5mm}
  \textbf{Nancy Kanwisher$^2$}
  \hspace{5mm}
  \textbf{Joshua B. Tenenbaum$^2$}
  \hspace{5mm}
  \textbf{Alexei A. Efros$^1$} 
  \\ \\
  $^{1}$University of California, Berkeley \hspace{2mm} $^{2}$Massachusetts Institute of Technology
}

\begin{document}

\maketitle

\vspace{-3mm}
\begin{figure}[h]
    \vspace{-5mm}
    \begin{center}
    \includegraphics[width=.9\textwidth]{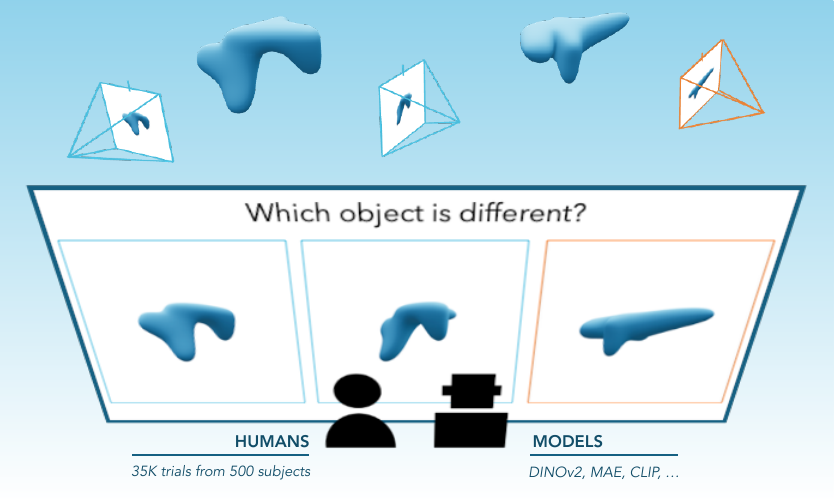}
    \end{center}
    \vspace{-1em}
    \caption{\small \textbf{How well do computer vision models represent the 3D structure of objects?} We develop a benchmark using a shape inference task from the cognitive sciences: Multiview Object Consistency in Humans and in Image models (MOCHI). Given three images of objects from random viewpoints, observers must identify which image depicts the object that is different. We compare human performance (35K trials from over 500 subjects, including accuracy, reaction time, and gaze data) against a number of standard computer vision models.}
    \label{teaser}
\end{figure}

\begin{abstract}
    We introduce a benchmark to directly evaluate the alignment between human observers and vision models on a 3D shape inference task. We leverage an experimental design from the cognitive sciences which requires zero-shot visual inferences about object shape: given a set of images, participants identify which contain the same/different objects, despite considerable viewpoint variation (example in Fig. \ref{teaser}). We draw from a diverse range of images that include common objects (e.g., chairs) as well as abstract shapes (i.e., procedurally generated `nonsense' objects). After constructing over 2000 unique image sets, we administer these tasks to human participants, collecting 35K trials of behavioral data from over 500 participants. This includes explicit choice behaviors as well as intermediate measures, such as reaction time and gaze data. We then evaluate the performance of common vision models (e.g., DINOv2, MAE, CLIP). We find that humans outperform all models by a wide margin. Using a multi-scale evaluation approach, we identify underlying similarities and differences between models and humans: while human-model performance is correlated, humans allocate more time/processing on challenging trials. All images, data, and code can be accessed via our \href{https://tzler.github.io/MOCHI/}{project page}.

\end{abstract}

\section{Introduction}

\vspace{-2mm}
Do computer vision models represent the 3D structure of objects? Answering this question is more difficult than it might appear. Many tasks that seem to require explicit 3D representations can be performed directly from 2D visual features. Depth estimation, for example, is commonly used to evaluate a model's 3D `awareness' (\cite{banani2024probing}), even though depth can be predicted using cues such as texture gradients (\cite{malik1997computing}), relative size (\cite{cutting1995perceiving}), shading (\cite{ramachandran1988perception}), and camera blur (\cite{mather1996image}). `3D understanding' may simply be a composite of such 2D features  (e.g., the Aspect Graph model; \cite{koenderink1979internal}), or it might require representations not directly accessible in these lower-level visual statistics (\cite{marr1982vision}). As large-scale vision models become more capable, these questions take on new implications. What methods might help us understand how vision models represent objects? 


\vspace{-.5mm}
The cognitive sciences have grappled with questions about 3D perception for decades. To understand human visual abilities, classic work in this field (\cite{shepard1971mental}) proposed a simple task: subjects were asked to determine whether two images contained the same or different objects, in spite of considerable viewpoint variation. Viewpoint variation was intended to prohibit low-level strategies (e.g., correspondence between 2D image features) and reveal veridical 3D shape understanding. Remarkably, the time needed to solve this task increases linearly with the angle of rotation between the two images, leading the authors to interpret these data as evidence for `mental rotation.' However, there are many non-3D strategies that might be equally effective (e.g., \cite{rock1987case, tarr1998image, bulthoff1992psychophysical, ullman1998three}). To evaluate these competing claims, the field has improved the experimental tasks used to probe human visual abilities (e.g., Fig. \ref{teaser}). 


\vspace{-.5mm}
How do computer vision models compare to humans on 3D tasks from the cognitive sciences? While humans are able to dramatically outperform contemporary vision models (\cite{bonnen2021ventral, oconnell2023approaching, bowers2022deep}), there's a catch. When images are presented briefly (e.g., $<$100ms), humans achieve roughly the same accuracy on 3D shape inferences as vision models; it is only when humans are given more time that we outperform these models (\cite{bonnen2023medial, ollikka2024humans}). Time is critical, in part, because it enables us to move our eyes around an image, sequentially attending to task-relevant features. Remarkably, when the neural structures responsible for integrating across visual sequences are damaged, human performance again resembles vision models (\cite{bonnen2021ventral}). These data suggest that 3D shape perception in humans is a dynamic process with temporally and neurally distinct stages; contemporary vision models simply capture the feedforward components of this process (\cite{jagadeesh2022texture, renninger2004scene}). 

\vspace{-.5mm}
Two aspects of these experiments are especially relevant for evaluating contemporary vision models. First, these tasks are  agnostic as to how shape information is represented, focusing instead on visual abilities. This is in contrast to current computer vision methods which make strong assumptions about how 3D representations are formatted (e.g., depth, point clouds). Second, human-model alignment provides a better guarantee that model performance might relate to more general notions of `3D understanding'. That is, models that best predict human behaviors are those models most likely to exhibit similar choice behaviors in related tasks (\cite{cao2021explanatory}). This requires going beyond determining if models achieve human-\textit{level} performance (i.e., average accuracy) and evaluating whether models exhibit human-\textit{like} performance (e.g., correlated choice behaviors). 

\vspace{-.5mm}
In this paper, we construct a benchmark to evaluate the correspondence between humans and vision models using experimental tasks from the cognitive sciences. We integrate traditional experiments characterizing the object-level visual inferences of humans with recent benchmarking approaches (e.g., \cite{rajalingham2018large}). The human behavior in this dataset includes explicit measures, such as choice behavior, as well as intermediate measures, such as reaction time and gaze patterns. We develop an approach to evaluate the performance of several standard computer vision methods (e.g., DINOv2, MAE, CLIP), and then compare humans and models. Critically, instead of evaluating human-model alignment using a single measure, we use a series of increasingly granular metrics. Coarse-grained human-model comparisons (e.g., average performance across all trials) enables us to evaluate whether models achieve human-level shape inferences, while more fine-grained metrics (e.g., patterns of choice behaviors or attention) enable us to determine how human-like models are.

\vspace{-2mm}
\section{Experimental design, data collection, and model evaluation methods}

\vspace{-3mm}
Here we outline the task used to probe 3D vision, images in this benchmark, and human data collection. Finally, we summarize the evaluation methods we use to determine model performance. 

\subsection{Behavioral task}

Our experimental design requires zero-shot visual inference about object shape: given three images, participants identify which contain the same/different objects. This design enables us to parametrically vary trial difficulty by changing the relative similarity between different objects (i.e., between object A and object B), and the viewpoint variation between images of the same object (i.e., between object A and the same object from a different viewpoint, A$'$). Moreover, this task imposes minimal verbal demands, enabling us to focus on perceptual processes and not language or semantic knowledge. We employ two variants of this task. In the `odd-one-out' task (\cite{bussey2002organization}) participants must identify the image within a triplet that contains an object that is different from the other two (i.e., select B given A, A$'$, and B). In the `match-to-sample' task, participants are presented a `sample' image (A), and then must select the `match' image (A$'$) and not a lure (B). 

\begin{figure}[ht!]
\begin{center}
\includegraphics[width=1\textwidth]{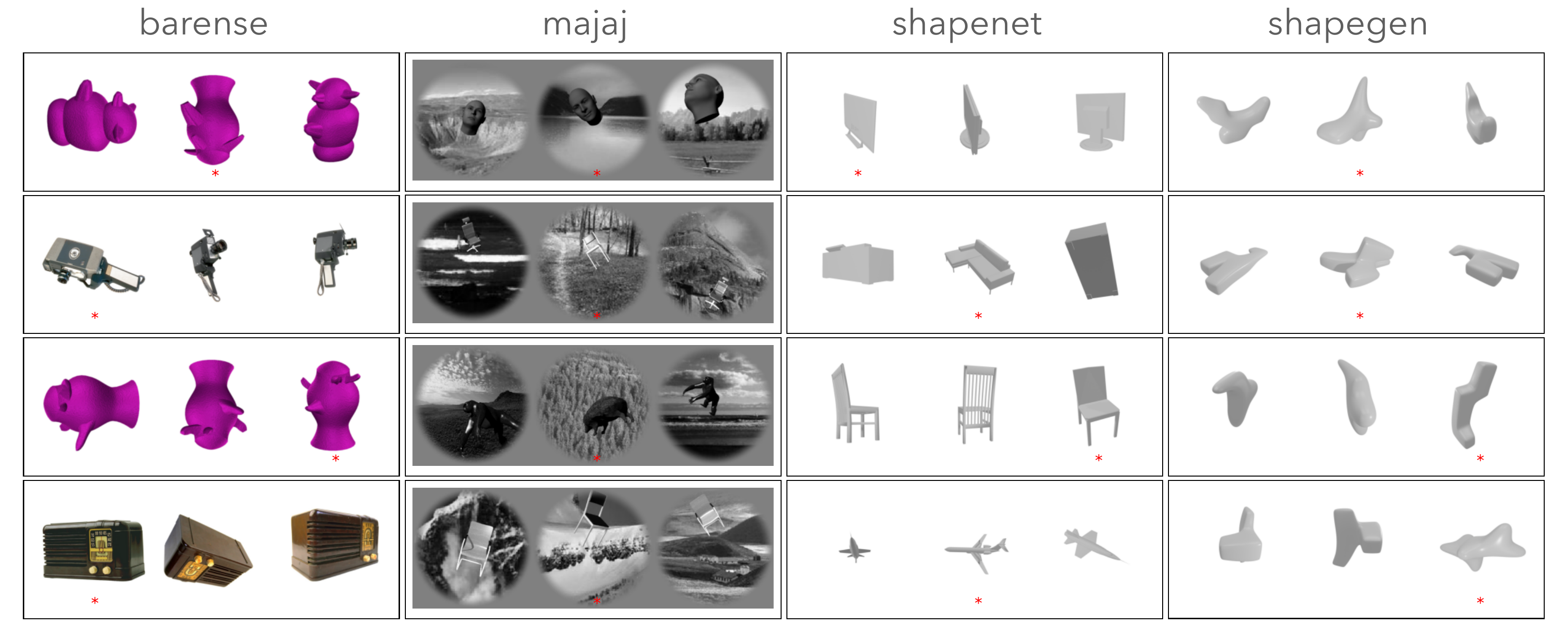}
\end{center}
\caption{\textbf{Example stimuli from the four datasets in MOCHI.} Each trial is composed of a triplet of images containing two objects: one from two different viewpoints (A and A$'$), and another object (B). Depending on the experiment, participants either infer the matching/non-matching object (pairing A-A$'$, or identifying B). B in each trial is marked with a \textcolor{red}{*} for illustrative purposes. Descriptions and examples of all categories in this benchmark can be found in the appendix (\ref{descriptions_of_categories}).} 
\label{example_triplets}
\end{figure}

\subsection{Stimuli}

We integrate experimental stimuli from four datasets collected by \cite{bonnen2021ventral} and \cite{oconnell2023approaching}. These images contain diverse object types and perceptual demands, which we organize into four distinct datasets. First, 'barense' contains color photographs of real-world objects (e.g., chairs, tables) and abstract shapes (i.e., synthetic objects without semantic attributes) on a white background (Fig. \ref{example_triplets} far left). These stimuli are separated into `high similarity' and `low similarity' conditions; `high similarity' trials are thought to rely on understanding the `compositional' 3D structure of objects while `low similarity' trials are thought to rely on simpler visual features (e.g., color, texture). Images from 'majaj' contain four categories of objects (animals, chairs, planes, and faces) rendered in  black and white and are superimposed onto randomized backgrounds (e.g., a chair floating in the sky above a mountain range; Fig. \ref{example_triplets} middle left). These stimuli are designed to make object segmentation and representation challenging, given the object-irrelevant distractors. We describe these as `barense' and 'majaj' according to the original source of these images (\cite{barense2007human, majaj2015simple}). Images from 'shapenet' contain everyday objects from multiple classes (e.g. cabinet, car, lamp) rendered without surface textures from random views sampled from a sphere  (Fig. \ref{example_triplets} middle right). The two objects selected for a given trial were always drawn from the same class. Images from `shapegen' contain procedurally-generated objects using the ShapeGenerator extension for Blender (Fig. \ref{example_triplets} far right). The result is a dataset which allows us to procedurally target a range of human behaviors when making zero-shot object-level inferences. None of these stimuli contain any personal identifying information or offensive content. We license all assets under CC BY-NC-SA 4.0.

\subsection{Human data collection}


After constructing over 2000 unique image triplets, we present these tasks to human participants, collecting 35K trials of behavioral data from over 500 participants via online and in-lab studies. Experimental data were collected online using Prolific. Participants were each paid \$15/hr for participating, and were free to terminate the experiment at any time. Experiments were approved by the MIT Committee on the Use of Humans as Experimental Subjects (Protocol \#1011004131). Experiments consisted of an initial set of instructions, 6 practice trials with feedback, and 150 main trials with no feedback. The 150 main trials were constructed such that no objects were repeated across trials, to avoid learning effects, and the ordering of the correct choice was randomized, to control for ordering effects. Given the performance we estimate for each image triplet, we normalize human accuracy to lie between zero (chance) and 1 (ceiling). This enables us to compare odd-one-out tasks and match-to-sample tasks in the same metric space. We collect eyetracking data on a subset of images and outline the data collection and analysis procedures for these data in the appendix.

\begin{figure}[ht!]
\begin{center}
\includegraphics[width=1\textwidth]{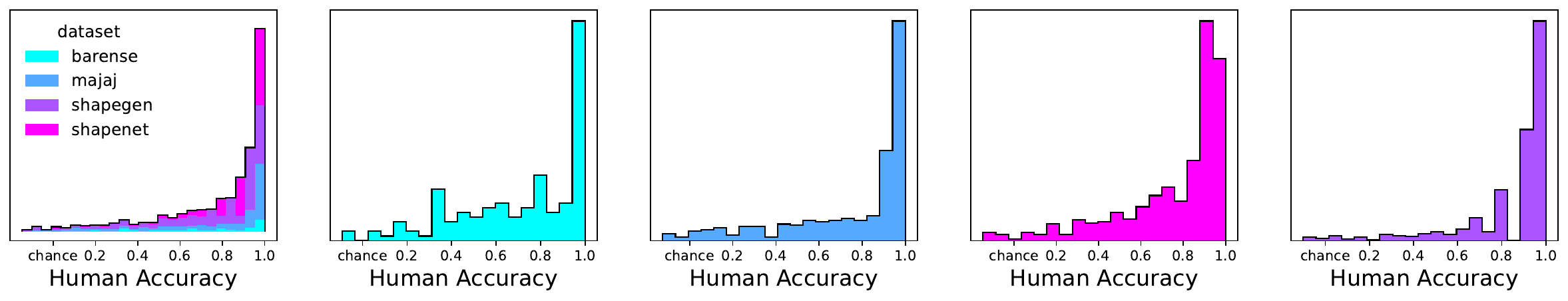}
\end{center}
\caption{\textbf{Distribution of human accuracy across trials in each dataset in this benchmark.} While humans are reliably accurate across trials, there is a long-tailed distribution of performance. This is by design, as it provides more challenging behavioral targets to model.}
\label{human_reliability_histograms}
\end{figure}

\subsection{Establishing a suitable model evaluation method for this benchmark}

We evaluate the performance of vision models optimized via contrastive (DINOv2) and autoencoding (MAE) self-supervision objectives. Furthermore, we analyse the performance of vision-language models trained via a constrastive image-text objective (CLIP) and with visual instruction tuning (LLaVA). Given this representative subset of available models, we evaluate multiple instance of each model class (e.g., DINOv2-Base, DINOv2-Huge, and DINOv2-Giant) in order to determine how model scale relates to performance on this benchmark as well as human-model alignment. Given an image triplet, composed of two images of object A from different viewpoints (i.e., A and A$'$) and a third image of object B, we extract model responses independently for each image. For all DINOv2 models we use pooled features after the final hidden layer. For all CLIP models we extract features from the final layer of the vision encoder. For MAEs we extract features from the [CLS] token. To estimate model performance we use several evaluation metrics on each trial, given the feature vectors in response to A, A$'$, and B. We model all trials as odd-one-out tasks and determine model performance as the accuracy the model achieves in identifying the non-matching object (B). 

\subsubsection{\textbf{Distance metrics}} 

We define an analytic approach that generates an estimate of the non-matching object in each trial: given model responses to A, A$'$, and B, we compute the pairwise similarity between items, then determine the `oddity' to be the item with the lowest off-diagonal similarity to the other images. We estimate the similarity between items using multiple distance metrics (cityblock, cosine, euclidean, l1, l2, manhattan, seuclidean, correlation, minkowski, chebyshev, braycurtis, and canberra). A one-to-one comparison with human performance requires that our we have a continuous estimate of model performance on each trial (i.e., not a single 1 or 0). To achieve this, we determine model performance for each triplet in the following way: for each iteration, we apply random in-plane rotations to the original images, estimate model performance, and average performance on this image across all iterations. The in-plane rotations we apply to each images are drawn from the same distribution of rotations used to generate the stimuli presented to humans. Thus, for each triplet we have an estimate of each models' averaged zero-shot performance, as well as estimates of the variance in this trial (e.g., standard error of the mean, SEM, and standard deviation, STD). 

\begin{figure}[ht!]
\begin{center}
\includegraphics[width=1\textwidth]{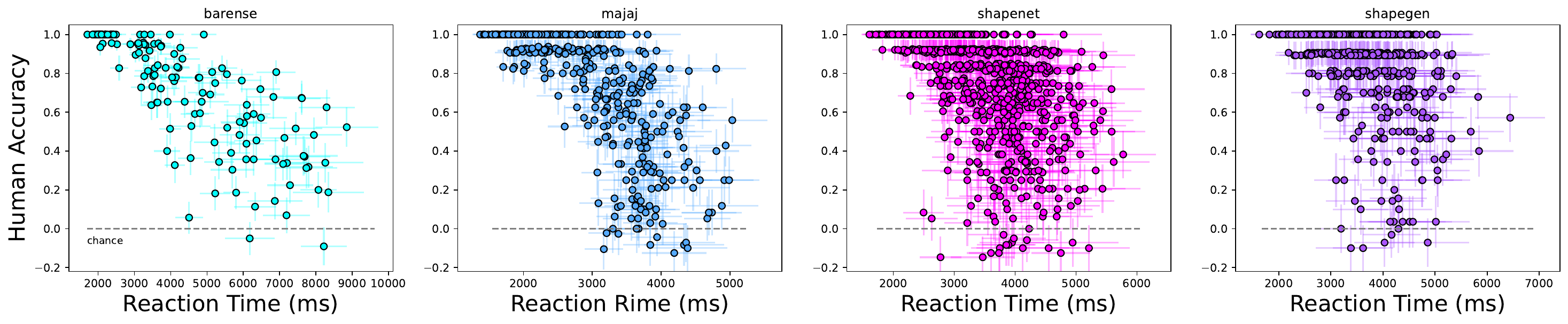}
\end{center}
\caption{\textbf{Accuracy and reaction time distributions for human participants across datasets.} Across datasets we observe a clear relationship between accuracy and processing time; as trials become more difficult, participants allocate more attention/time. Critically, the distribution of human behavior ranges from chance to ceiling, indicating that we have a suitable estimate of the full range of human visual abilities. This and all subsequent error bars are SEM computed over trials. }  
\label{accuracy_reaction_time_human}
\end{figure}

\subsubsection{\textbf{Linear probes}} 

We design two independent linear probes to evaluate model performance on each triplet. We first design a standard linear probing strategy following the default setting in~\cite{caron2021emerging, }. A trainable MLP layer is applied to a frozen visual backbone. We formulate this task as a multi-way classification problem. Additionally, we design a lightweight linear probe hand-tailored for our experimental procedure: for a given triplet, T, we train a linear classifier to perform a `same-different' classification on similar (not T) triplets in each dataset, and evaluate on triplet T. Concretely, in each condition (e.g., `planes' in the Majaj dataset), for each triplet, we compute pairwise difference vectors between all images (i.e., $A - A'$ = $\Delta_{A-A'}$) and label this vector appropriately (i.e., label($A - A'$) = 1,  label($A$ - $B$) = 0). This results in a series of difference vectors and their corresponding labels. Critically, this enables us to learn a same-different decision boundary that generalizes to different decisions within this same condition. To this end, for each triplet T, we train an SVM to perform this classification using a subset of triplets from the same condition (e.g., 75\% of `planes' discriminations if T is in the `planes' condition) and evaluate on trial T. We repeat this procedure 100 times for each trial, resulting in a continuous measure of model performance include SEM and STD.

\begin{figure}[ht!]
\begin{center}
\includegraphics[width=1\textwidth]{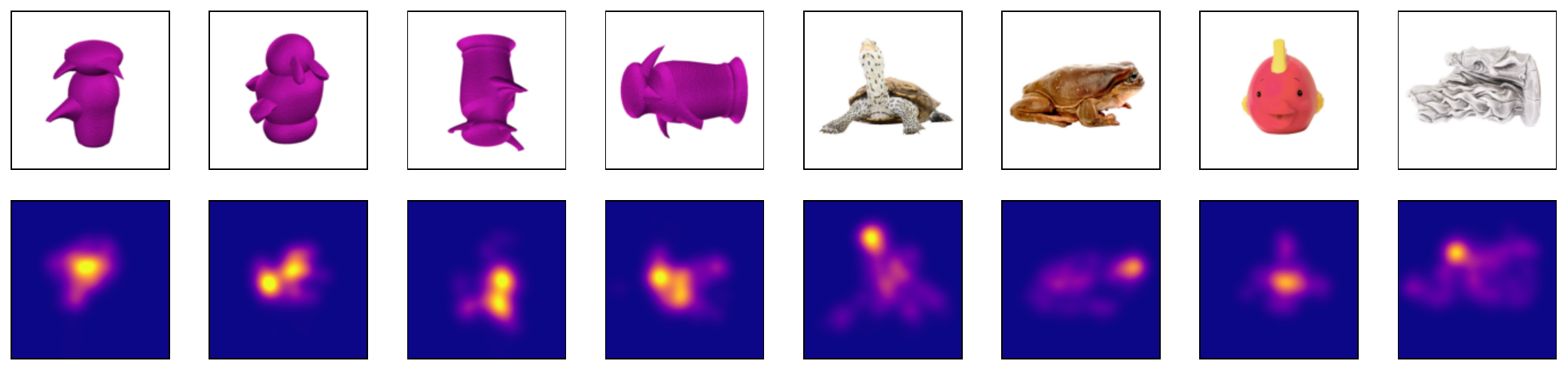}
\end{center}
\caption{\textbf{Examples of human saliency maps collected on a subset of images in MOCHI.} Given the foveal nature of primate vision, humans must move their gaze in order to collect high-acuity visual information. As such, measuring human gaze patterns reveals human attention patterns. We collect gaze behavior from human participants on all trials in one stimulus set (barense).
}
\label{human_salience_examples}
\end{figure}

\subsubsection{\textbf{A lightweight linear probe exhibits the most promising performance on this benchmark}}

Distance metrics and linear probes each have desirable properties; distance metrics provide a true `zero-shot' estimate of a model's representations, while linear probes offer a more expressive readout tailored to this specific task. Both have undesirable properties as well; distance metrics typically assume that task-relevant information is uniformly represented in model features, while linear probes require task-relevant training data and decisions about numerous design choices (e.g., hyperparameters of the linear probe and defining a suitable train-test split). Here we simply choose the evaluation approach that performs best. We find that our same-different classifier has a clear advantage of more standard linear probes and distance metrics. Training a standard linear probe is unstable, leading to catastrophic performance. This approach overfits quickly and fails to learn useful representation under this setting, perhaps due to the relatively small dataset size. Distance metrics exhibit much better performance without any specific fine-tuning. However, the same-different linear classifier performs significantly better than both the average performance of the distance metrics (paired t-test $t(2018)$ = $12.83, P = 2.74$x$10^{-36}$) as well as all individual metrics (e.g., $l2$, paired t-test $t(2018)$ = $12.40, P = 4.3$x$10^{-34}$). We visualize this comparison across datasets in Fig. \ref{comparing_distance_metrics_scatter}. For subsequent analyses we report the results of model performance using the lightweight same-different classifier. As a final step, we normalize the accuracy to be between 0 (chance) and 1 (ceiling). 

\section{Results}

\subsection{Human accuracy, reaction time, and gaze behaviors are reliable across participants.} 

Humans behaviors are both accurate and reliable, exhibiting considerable variation across conditions in this benchmark. Average human accuracy is 78\%, with performance ranging between 40\% and 100\% for different conditions. To determine the noise ceiling for these behaviors---i.e., how much variance we might hope to explain with vision models---we compute the split-half reliability of these behaviors over 1000 permutations and compare this distribution to an empirically estimated null (i.e., generated by shuffling the order of the split-half correlation). We find that human performance is reliable at the level of averaged performance across datasets ($r_{median}$=.93, paired t-test $t(999)$ = $49.1, P = 2.24$x$10^{-268}$), averaged performance across conditions ($r_{median}$=.95,  paired t-test $t(999)$ = $48.993,  P = 6.7$x$10^{-268}$), and averaged performance across trials  ($r_{median}$=.93, paired t-test $t(999)$ = $958.17, P = 0$). Beyond performance itself, we observe a significant relationship between accuracy and reaction time across all trials in this dataset; as accuracy decreases, reaction time increases (r=-.52, $F(1, 2017)$ = $-27.44, P = 4 $ x $ 10 ^{-141}$). This suggests that more difficult trials require more processing time for relatively accurate performance. Reaction time is also reliable across datasets   ($r_{median}$=.99, paired t-test $t(999)$ = $52.68, P = 1.27 $ x $ 10 ^{-290}$), across conditions ($r_{median}$=.98, paired t-test $t(999)$ = $152.9, P = 0$) and across trials ($r_{median}$=.69, paired t-test $t(999)$ = $903.4, P = 0$). Finally, we turn our attention to the most fine-grained behavioral measure we collected on a subset of the experimental data (all stimuli in barense).  We estimate the split-half reliability of human gaze patterns by determining how correlated  salience maps are across participants who viewed the same image. These measures are reliable across participants  ($r_{median}$=.94, paired t-test $t(83998)$ = $430.9, P = 0$, Fig. \ref{human_gaze_reliability}). That is, for a given image, different people attend to similar object locations/features. Taken together, these results suggest that while humans are highly accurate, their performance exhibits considerable variance. Critically, these behavioral data are reliable across participants, suggesting that there is variance to explain at multiple levels of granularity. 

\begin{figure}[ht!]
\begin{center}
\includegraphics[width=1\textwidth]{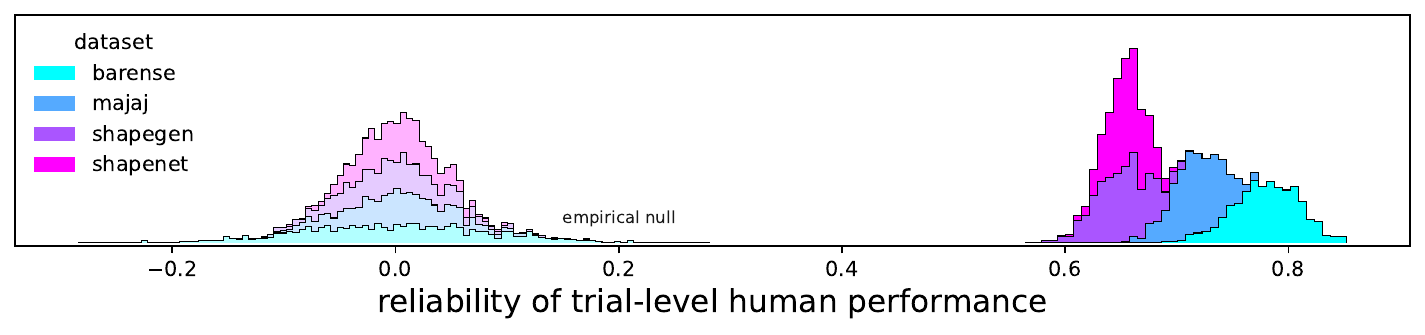}
\end{center}
\caption{\textbf{Distribution of reliability scores for human performance.} The data in this benchmark represent consistent, reliable properties of human object-level shape inference, as evidenced by the high split-half reliability (right), contrasted with the shuffled, empirical null (left), 
}
\label{dataset_histograms}
\end{figure}

\subsection{Humans outperform vision models by a wide margin, regardless of model scale}

\begin{figure}[ht!]
\begin{center}
\includegraphics[width=.7\textwidth]{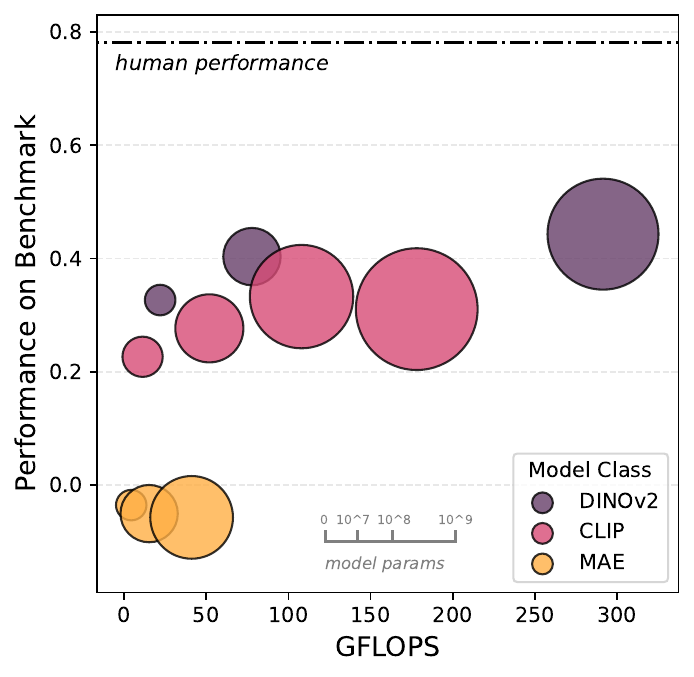}
\end{center}
\caption{\textbf{Relationship between MOCHI performance and model size/FLOPS.} Model performance improves with increased scale for some model types (e.g., DINOv2, purple) and not others (MAE, yellow). Nonetheless, humans (dashed line) outperform all models by a wide margin. 
}
\label{model_scale}
\end{figure}

Humans outperform models on this benchmark by a wide margin. The best performing model (DINOv2-G) achieves 44\% accuracy while humans achieve 78\%. On average, DINOv2 models achieve an average accuracy of 0.43\%, while CLIPs have an average accuracy of 0.29\%, and MAEs have an average accuracy of -0.03\%. DINOv2 outperforms all other models on this benchmark (Fig. \ref{model_scale}, all DINOv2s vs all CLIPs, paired t-test $t(16150)$ = $14.31, P = 3.17 $ x $ 10 ^{-46}$, vs all MAEs, paired t-test $t(12112)$ = $47.23, P = 0$). Conversely, MAE's performance hovers around chance for all model sizes ($t(6056)$ = $.11, P = 0.91$). Increased model size leads to improved performance on this benchmark for both CLIP and DINOv2, while the performance of MAEs does not improve. Insofar as object-level shape inferences are concerned, not all self-supervision objectives benefit from increased model scale. Moreover, a thirteen-fold increase in the number of model parameters from DINOv2-Base to DINOv2-Giant only leads to an increase from 32\% to 44\%.





\begin{table}[ht!]
    \resizebox{\linewidth}{!}{
        \begin{tabular}{lrrrrrrrrrr}
        \toprule
         & Human &DINOv2-G & DINOv2-B & DINOv2-L & CLIP-B & CLIP-L & CLIP-H & MAE-B & MAE-L & MAE-H \\
        \midrule
        Accuracy & \textbf{0.78} & 0.44 & 0.33 & 0.40 & 0.23 & 0.28 & 0.33 & -0.04 & -0.05 & -0.06 \\
        \bottomrule
        \end{tabular}
    }
    \vspace{1em}
    \caption{Performance of humans and vision models on our MOCHI benchmark. Absolute accuracy is normalized in relation to chance (zero) and ceiling (one).} 
\end{table} 

\subsection{While humans outperform models, their performance is correlated. }

\begin{figure}[ht!]
\begin{center}
\includegraphics[width=1\textwidth]{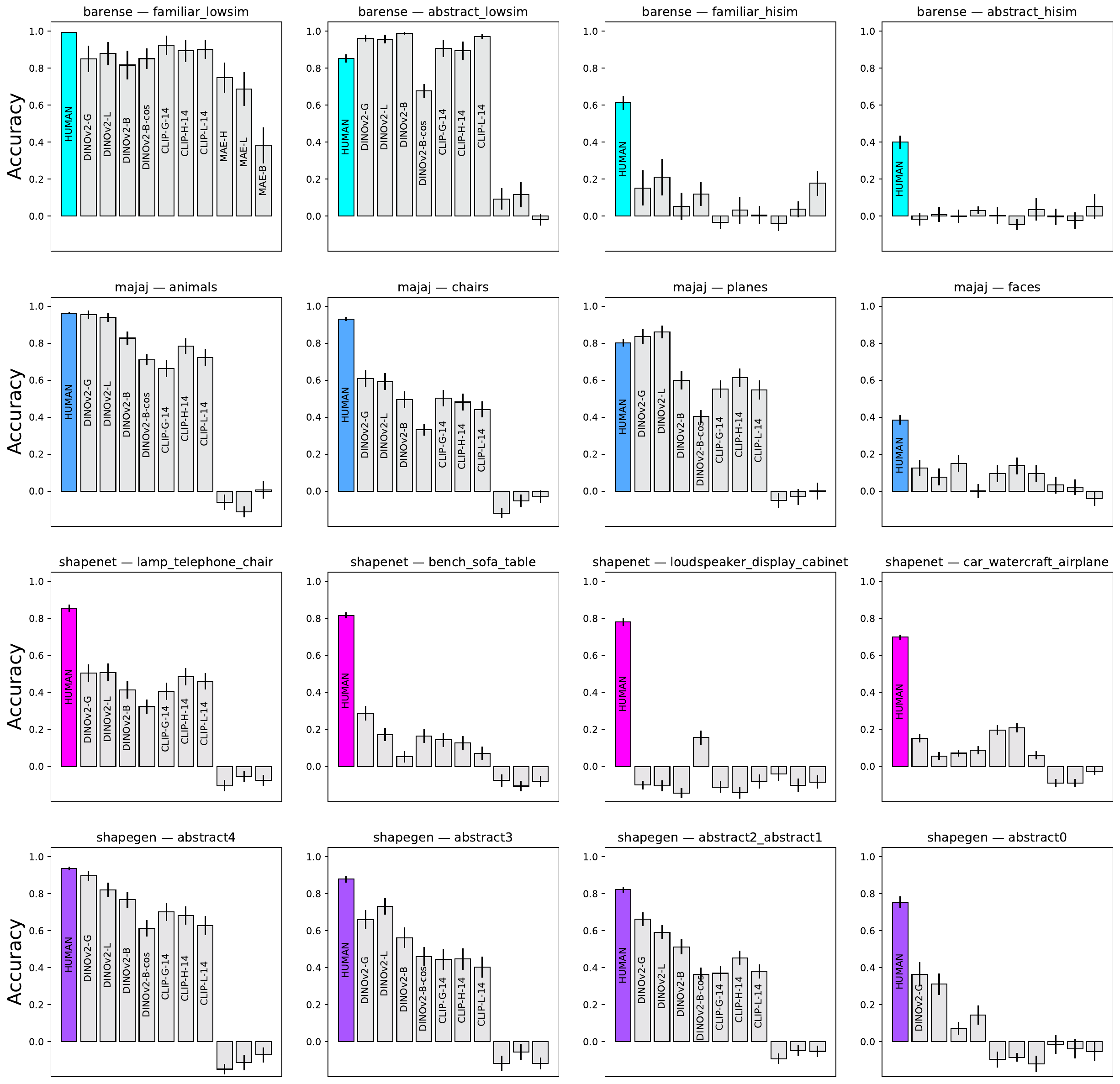}
\end{center}
\caption{\textbf{Mean difference between humans and multiple vision models across conditions.} For each condition in this dataset, we compare human performance (left, in color) to DINOv2 models, CLIPs, and MAEs across multiple scales (greys).}
\label{human_model_condition}
\end{figure}

\begin{figure}[ht!]
\begin{center}
\includegraphics[width=1\textwidth]{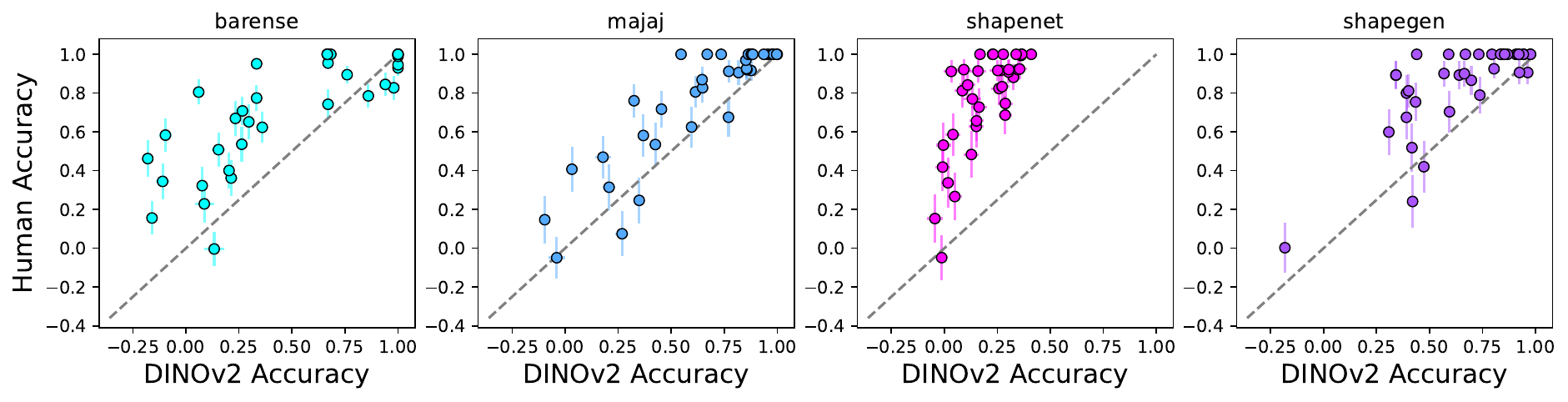}

\caption{\textbf{Human and model performance is correlated.} Humans outperform vision models on the MOCHI benchmark. Nonetheless, there is a correlation between human and model performance. We visualize this relationship by binning across trials and plotting the performance of DINOv2-G along the x-axis and human performance along the y-axis, for each dataset in this benchmark.
}
\label{human_model_accuracy_correlation}
\end{center}
\end{figure}

When we move beyond averaged accuracy of humans and models, we find that human and model performance is correlated. We begin by noting that the gap between human and model performance varies considerably across datasets (average difference 38\% $\pm$ 30\% STD). We first report some qualitative patterns observed across different conditions, in relation to their stimulus properties. The best performing model we evaluated (DINOv2) approaches human-level performance on a relatively diverse subset of experimental conditions: in the `animals' and `planes' conditions in the majaj dataset, which have greyscaled objects superimposed onto random backgrounds (Fig. \ref{human_model_condition}, second row), `abstract4' in the shapegen dataset, which has abstract objects removed of color- or texture-level visual identifiers (Fig. \ref{human_model_condition}, fourth row, left), and in `familiar\_lowsim' and `novel\_lowsim', two conditions in barense (Fig. \ref{human_model_condition}, top left). Nonetheless, there many more conditions where model performance drops to chance as human performance is relatively intact (e.g., `familiar\_hisim' in barense, Fig. \ref{human_model_condition}, top row, and `abstract0' in shapegen, bottom row). Returning to quantitative measures, we find that while there is a considerable performance gap between humans and DINOv2, their behaviors are nonetheless correlated. We observe this human-model correlation across datasets ($r=.42$, not reporting statistics because of low sample size, n=4), conditions ($r=.58$, $F(1, 23)$ = $3.40, P = 0.002 $), and trials ($r=.35$ $F(1, 2017)$ = $16.86, P = 9 $ x $ 10 ^{-60} $). When comparing to the reliability ceiling estimated for human behaviors across these different resolutions, DINOv2 can predict about 0.61\%  and 0.37\% of the explainable variance across conditions, and trials, respectively (Table \ref{multiple_resolution_correlation}) . We visualize this relation in Fig. \ref{human_model_accuracy_correlation}, binning across trials. Taken together, these data suggest that task difficulty is shared across humans and models, despite considerable differences in accuracy. 




\begin{table}[ht!]

\resizebox{\linewidth}{!}{
    \begin{tabular}{lrrrrrrrrr}
    \toprule
    Level & DINOv2-B & DINOv2-L & DINOv2-G & CLIP-B & CLIP-L & CLIP-H & MAE-B & MAE-L & MAE-H \\
    \midrule
    Condition & 0.52\% & \textbf{ 0.60\%} & \textbf{0.60\% }& 0.56\% & 0.54\% & 0.53\% & -0.09\% & 0.12\% & 0.08\% \\
    Trial & 0.19\% & 0.18\% & \textbf{ 0.28\%} & 0.18\% & 0.14\% & 0.22\% & 0.02\% & 0.01\% & 0.06\% \\   
    \bottomrule
    \end{tabular}    
}
\vspace{1em}
\caption{\textbf{Alignment between human and model performance across levels of granularity. } We determine the split-half correlation between humans and vision models at the condition and trial level, and scale this value by the median split-half reliability of human behavior at this resolution. 
}
\label{multiple_resolution_correlation}
\end{table} 

\subsection{Human-model divergence may be explained by increased processing time and attention}

To gain further insights into the superior performance of humans compared to vision models, we turn to additional measures of human behavior: reaction time and gaze dynamics (i.e., attention). Reaction time is commonly used as a measure for the amount of processing needed for a given behavior, and gaze behaviors are a direct measure of the visual features that humans attend to when viewing an image. These behavioral intermediates provide clues about the algorithmic basis of human visual abilities. First, we directly compare the relationship between human reaction time and model performance. We find that the trials for which model performance is degraded are the same trials where humans allocate more time (i.e., a significant correlation between human reaction time and model performance, r=-0.29, $F(1, 2017)$ = $-13.72, P = 5 $ x $ 10 ^{-41}$, Fig. \ref{human_model_accuracy_correlation}). Given that human performance on these trials is significantly greater than model performance (i.e., significant above-diagonal variance observed in Fig. \ref{human_model_accuracy_correlation}), it appears that humans allocate more processing time for those trials that are more challenging. If this were true, we might expect this processing time to be evident in some way in their attention patterns. Specifically, it has been hypothesized that human viewing behaviors (i.e., sequentially sampling visual images by moving our gaze to different visual features) enables us to `compose' flexible representations of objects using a finite set of visual representations (\cite{ullman1987visual, bonnen2023medial}). As we outlined in the human behavioral results, these gaze dynamics are highly reliable across participants; when two people encode a visual stimulus, they tend to look at the same features. Are vision models attending to the same feature that humans are, during these extending viewing periods? We extract attention measures from intermediate model layers, across all models layers, and we find that DINOv2 features do not predict human visual attention any better than would be expected from chance (human-human reliability scores visualized in Fig. \ref{human_gaze_reliability}, left, alongside model-human attention scores; random subset of attention maps shown in Fig. \ref{human_gaze_reliability} right). The patterns observed in human attention are not evident in model attention. 



\begin{figure}[ht!]
\begin{center}
\includegraphics[width=1\textwidth]{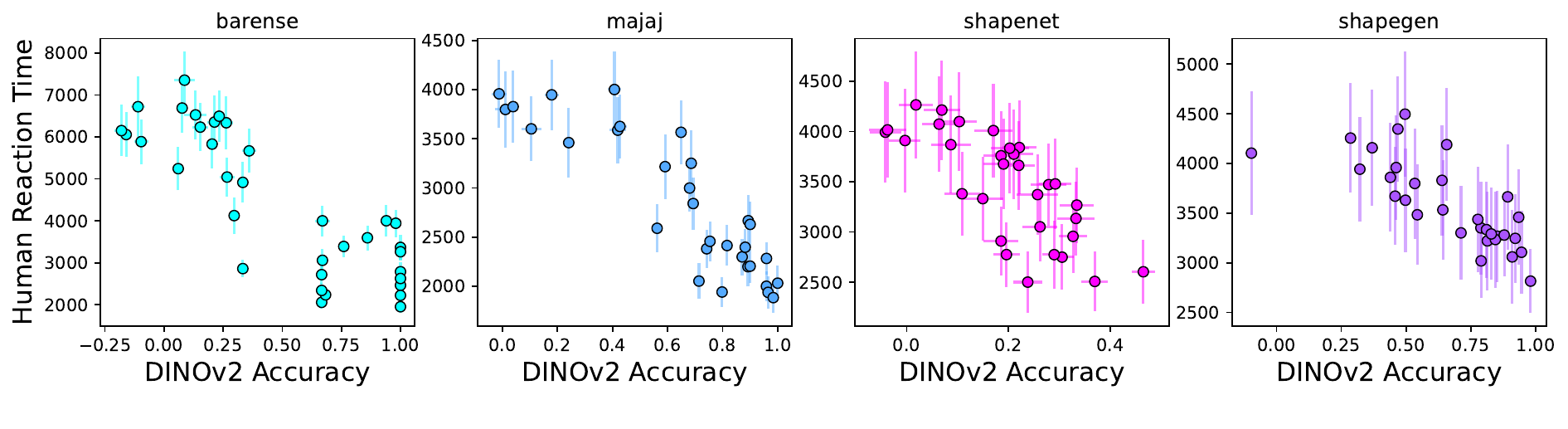}
\caption{\textbf{Humans spend more time on trials where models fail.} We visualize the relationship between binned model performance of DINOv2-G (x-axis) and binned human performance (y-axis) across all four datasets. Humans spend more time on trials for which model performance degrades.}
\label{human_model_accuracy_rt_correlation}
\end{center}
\end{figure}

\begin{figure}[ht!]
\begin{center}
\vspace{-1em}
\includegraphics[width=1\textwidth]{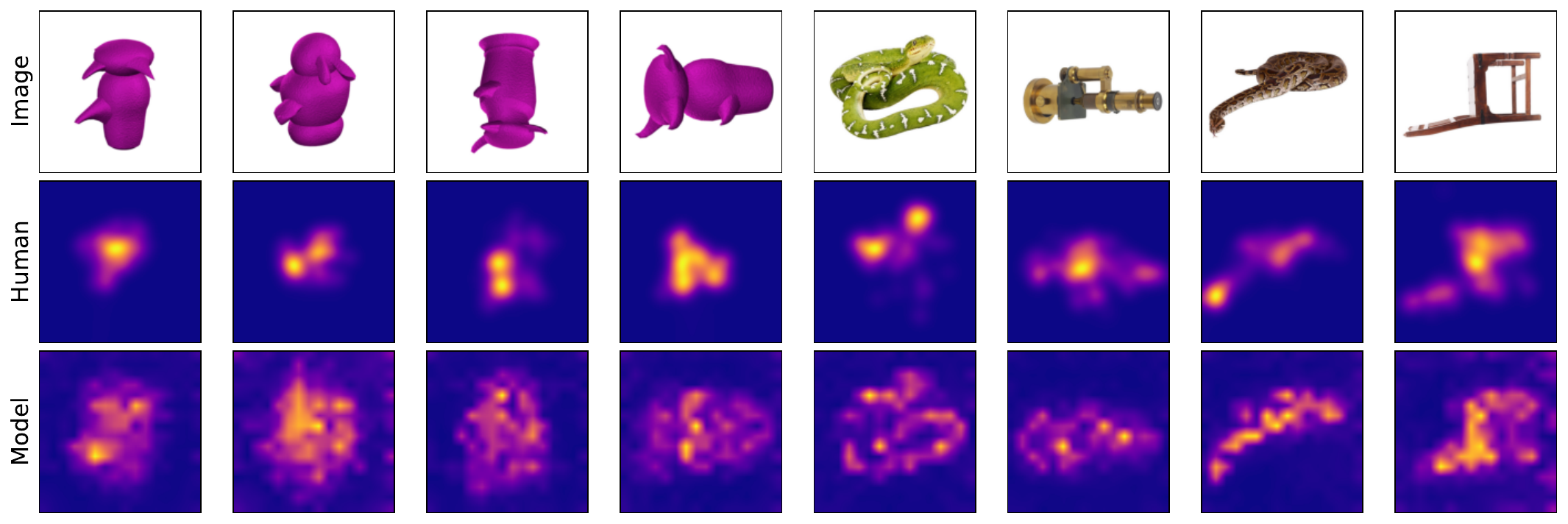}
\caption{\textbf{Humans reliably attend to object features not captured by DINOv2 attention maps.} To better understand the temporal dynamics that enable humans to outperform models, we collect gaze data on a subset of trials in this benchmark. While humans are focused on task-relevant features, model attention is distributed across the object more uniformly. Critically, these gaze patterns in humans are reliable across participants, suggesting that there is meaningful variance to explain. }
\label{model_attention_visualization}
\end{center}
\end{figure}

\section{Conclusion}

We provide a benchmark to evaluate the alignment between human participants and vision models on object-level shape inferences. We develop readout methods to estimate model performance on these tasks, as well as multi-level evaluation metrics to compare models and human observers. When comparing average performance, humans outperform all vision models by a wide margin. Fine-grained behavioral measures provide clues about the nature of this human-model divergence. In spite of the mean difference between humans and models, their performance is correlated. This correspondence is profoundly asymmetric: when human accuracy decreases, model performance often falls to chance levels, while human accuracy is relatively robust on those trials where model performance degrades. This is evident when comparing models to intermediate behavioral measures relating to processing (reaction time) and attention (gaze patterns). Those trials where model performance degrades are the same trials that humans spend more time to resolve. Similarly, there is reliable variance in the human attention dynamics, as measured via eyetracking, which is unexplained by model attention. This benchmark identifies unambiguous failure modes in contemporary vision models, while providing clues as to the algorithmic basis of human visual abilities. This benchmark is designed to serve as an independent validation set, showcasing the utility of a multiscale approach to evaluating human-model alignment on object-level shape inferences.

\section{Data and code availability}

We license all assets (data, code, and images) under CC BY-NC-SA 4.0. These resources can be found at our project page (\href{https://tzler.github.io/MOCHI/}{https://tzler.github.io/MOCHI/}). Images can be downloaded as a huggingface dataset (\href{https://huggingface.co/datasets/tzler/MOCHI}{https://huggingface.co/datasets/tzler/MOCHI}). Code and all benchmark results, from humans and all models evaluated, is available on github (\href{https://github.com/tzler/mochi_code}{https://github.com/tzler/mochi\_code}). 

\section{Acknowledgements}

We thank Devin Guillory for suggestions about initiating this project and Sophia Koepke for feedback on this manuscript. This work is supported by the National Institute of Neurological Disorders and Stroke of the National Institutes of Health (Award Number F99NS125816) as well as the National Science Foundation (Grant 2124136).

\printbibliography


\appendix

\section{Appendix / supplemental material}

\subsection{Online human data collection}

Human experimental data were collected online via Amazon Mechanical Turk and Prolific via experiments were implemented in JsPsych (\cite{de2015jspsych}). Each experiment began with an instruction phase, which introduced them to the task as well as provided 5 practice trials. This provided an opportunity for participants to acclimate themselves to the task and the controls. Once the experiment began, participants initiated the beginning of each trial with a button press (spacebar), such that they can (effectively) pause the experiment whenever they deem appropriate. This was designed to reduce environmental interference in the experiment. Experiments were designed to be completed in 10 minutes and participants were payed at a rate of roughly \$16/hour. In addition, participants were awarded a bonus commensurate with their performance, enabling them to earn up to twice the base pay. In order to ensure that participants were fairly compensated for their time, even in the case of a crowdsourcing platform errors, trial-by-trial data were collected throughout the experiment and stored on a custom server built from a Digital Ocean `droplet.' 

We administer two related experimental designs. First, we use a 3-way concurrent visual discrimination task commonly used to evaluate the role of MTC in perception (\cite{buckley2001selective, bussey2002perirhinal, barense2007human}). This design enables us to determine visual inferences that are possible with unlimited viewing time, as all stimuli remain on the screen for the duration of the trial. On each trial, participants are presented with three images and must identify the image that does not match the other two in terms of object identity (i.e., the `oddity'). Participants are given upwards of ten seconds to complete each trial. At any point in this duration, participants can select the oddity with a button press (right arrow, left arrow, or down arrow) corresponding to those locations on the oddity array. After this button press, participants are given feedback related to their performance on that trial, indicating whether their choice was correct or incorrect. If participants do no press a button in these ten seconds, the trial is marked as incorrect, feedback is given on the screen encouraging them to complete each trial within the allotted time. 

\subsection{Eye tracking data collection}

Eye tracking was performed using an infrared video-based eye-tracker at 1000 Hz (Eyelink 1000; SR Research). Stimuli were displayed on a 22.5 inch VIEWPixx LCD display (resolution of 1900×1200, refresh rate of 120 Hz) and responses collected via keyboard. Other sources of light were minimized during data collection. The stimulus on the sample screen was presented at the central field of view and spanned up to 10 degrees of visual angle. This stimulus size was selected such that in order to collect high-acuity visual information from various stimulus locations, participants had to move their eyes (i.e., make a saccade). Stimuli on the match screen were the same size, but presented side by side, offset from the horizontal midpoint of the screen by 10 degrees of visual angle. Each experiment began with gaze calibration, then 5 practice trials to acclimate participants to the experimental setup. Each trial was initiated by the participants and began with participants maintaining fixation at the center of the screen (to perform drift correction at the beginning of each trial). Participants completed each trial at their own pace and there was a brief rest period every 5 minutes. This duration of this rest period was at the discretion of each participant. After this rest period, there was another gaze calibration, after which participants again completed a series of trials at their own pace as described above. For all gaze analyses (e.g., evaluating gaze reliability) we estimate gaze-related events (e.g. fixations) directly from the raw gaze data using a standard python library (\hyperlink{https://github.com/psychoinformatics-de/remodnav}{REMoDNaV}; \cite{nystrom2010adaptive}). 

\subsection{Estimating gaze reliability} 

We estimate the split-half reliability of in-lab gaze dynamics using the following protocol. First, for each trial, a subject-level salience map is generated from the raw gaze behaviors: a 2D histogram is generated from the raw time series, which is then smoothed with a Gaussian kernel. We note that the results reported in this manuscript are robust to the resolution of the 2D histogram and size of the smoothing kernel. This protocol yields a salience map for each image for each subject. We then generate a random split of subjects and partition the salience maps for a given image using this random subject split. We then average across participants in each random split, which results in two salience maps, each corresponding to the random split of participants allotted to that half. We then  estimate the correlation between the two (random split-half) salience maps associated with this image. We repeat this protocol for 100 random split-half permutations (i.e., generating a new shuffle of participants each iteration). For each image, we then have a distribution of split-half correlations which enables us to evaluate how similar participants viewed each image. To establish an empirical null we compute the correlation between random splits corresponding the different images within the same trial. Additionally, we estimate the bottom-up salience of each image (\cite{itti1998model}) and compute the correlation between this bottom-up salience map and the random splits associated with each permutation of each image.

\begin{figure}[h!]
\begin{center}
\includegraphics[width=.8\textwidth]{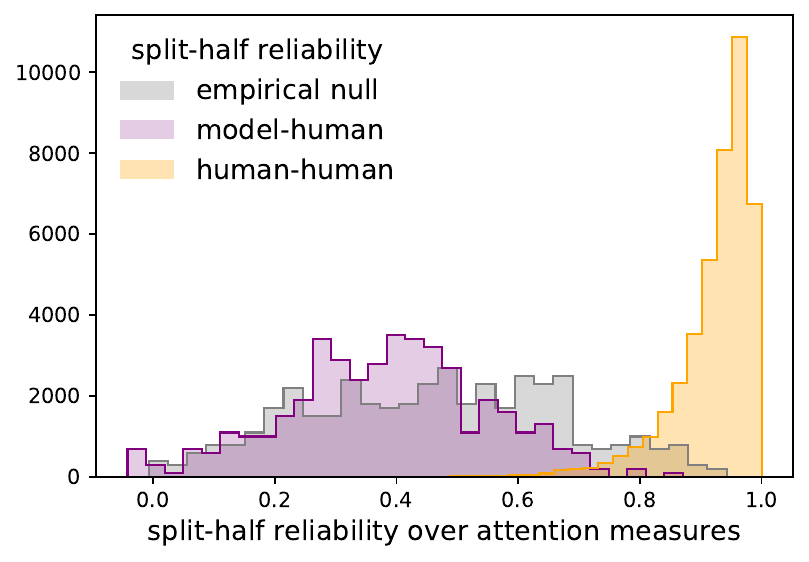}
\end{center}
\vspace{-2mm}
\caption{\textbf{Human gaze patterns are reliable but not predicted by model self-attention.} While many human behavioral measures are predicted by model performance, self-attention does not predict human attention, suggesting that encoding strategies for humans and machines are quite different. }
\vspace{-5mm}
\label{human_gaze_reliability}
\end{figure}




\begin{figure}[ht!]
\begin{center}
\includegraphics[width=.45\textwidth]{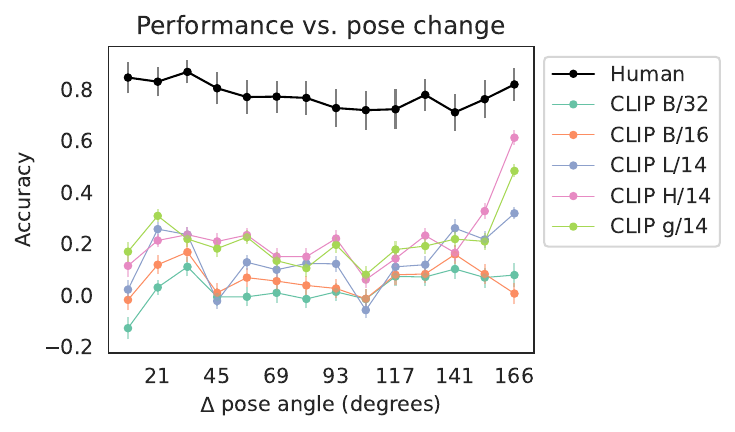}
\includegraphics[width=.45\textwidth]{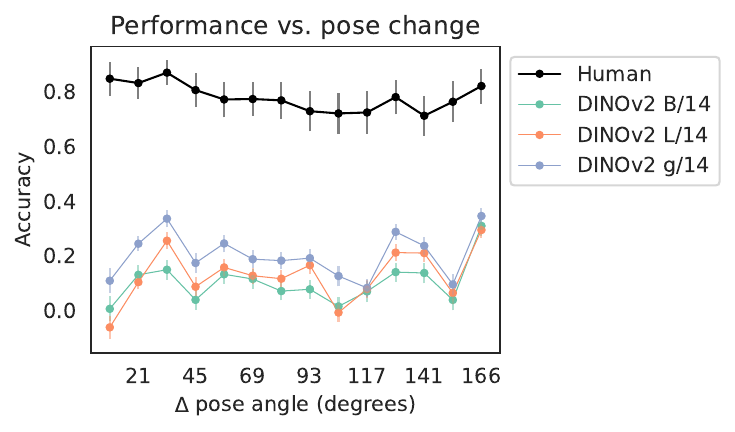}
\includegraphics[width=.45\textwidth]{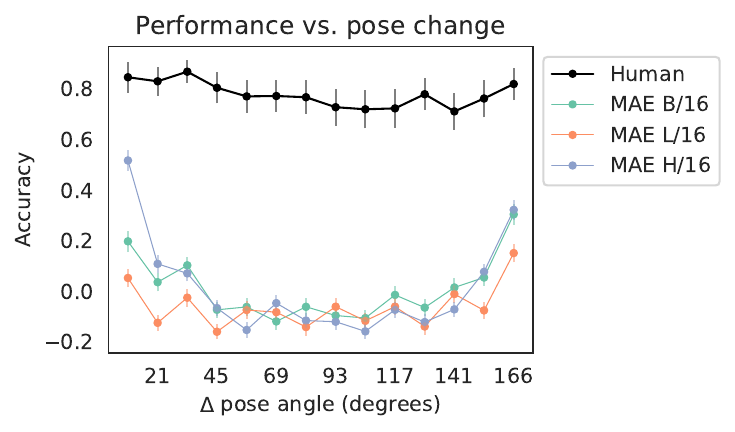}
\end{center}
\caption{\textbf{Visualizing viewpoint tolerance in humans and models across stimuli in shapenet}}
\label{dino_human_triallevel}
\end{figure}

\subsection{Stimuli} 

Data from \cite{barense2007human} include common, real-world objects (e.g., color photographs of chairs, tables) and abstract shapes (i.e., synthetic objects without semantic attributes) on a white background (Fig. \ref{example_triplets} far left). These stimuli are separated into `high similarity' and `low similarity' conditions; `high similarity' trials are thought to rely on understanding the `compositional' 3D structure of objects while `low similarity' trials are thought to rely on simpler visual features. Images from \cite{majaj2015simple} belong to four categories of objects: animals (e.g., gorillas, lions), chairs, planes, and faces. Images are in black and white, all from multiple viewpoints, and are superimposed onto randomized backgrounds (e.g., a chair floating in the sky above a mountain range; Fig. \ref{example_triplets} middle left). These stimuli are designed to make object segmentation and representation challenging, given the object-irrelevant distractors. The authors from \cite{barense2007human} and \cite{majaj2015simple} were contacted directly and provided explicit consent for using these stimuli for all future modeling work. Data from \cite{oconnell2023approaching} includes two synthetic datasets rendered in Blender from manmade (ShapeNet) and abstract (ShapeGen) objects. The ShapeNet dataset uses objects from multiple classes (e.g. cabinet, car, lamp) rendered with surface textures removed from random views sampled from a sphere around the object (Fig. \ref{example_triplets} middle right). The two objects selected for a given trial were always drawn from the same class. The ShapeGen dataset is composed of algorithmically-generated objects using the ShapeGenerator extension for Blender (Fig. \ref{example_triplets} far right). These objects are created by iteratively extruding random mesh faces from a base shape, and applying a Catmull-Clark modifier to produce smooth edges on the final objects. This pipeline can generate infinitely many unique shapes, while simultaneously controlling for the similarity of any two objects.  These objects were rendered from a circular yaw sweep with the camera pointed 45 degrees down at the object. The result is a dataset which allows us to procedurally target a range of human behavior when making zero-shot 3D shape inferences. 

\subsubsection{Description of all stimulus conditions} 

Each condition has its uniqe properties and belongs to one stimulus set (either barense, majaj, shapenet, or shapegen). Each stimulus set has properties that are shared across all conditions (e.g., all majaj conditions contain greyscale objects on randomized backgrounds, all shapegen conditions contain procedurally generated non-semantic objects).   
\begin{enumerate}

    \item \textbf{Familiar objects low similarity (barense)}: Objects from familiar categories (e.g., cars) in color with a white background; little similarity between matching/non-matching object. 
    \item \textbf{Familiar objects high similarity (barense)}: Objects from familiar categories (e.g., cars) in full color on a white background; high similarity between matching/non-matching object. 
    \item \textbf{Novel objects low similarity (barense)}: Semantically meaningless objects (greebles) in full color on a white background; low similarity between matching/non-matching object. 
    \item \textbf{Novel objects high similarity (barense)}: Semantically meaningless objects (greebles) in full color on a white background; high similarity between matching/non-matching object.
    \item \textbf{Animals (majaj)}: Eight different animals, greyscale, random category irrelevant background (e.g., mountains, beach). 
    \item \textbf{Chairs (majaj)}: Eight different chairs, greyscale, random category irrelevant background (e.g., mountains, beach). 
    \item \textbf{Planes (majaj)}: Eight different planes, greyscale, random category irrelevant background (e.g., mountains, beach). 
    \item \textbf{Faces (majaj)}: Eight different faces, greyscale, random category irrelevant background (e.g., mountains, beach). 
    \item \textbf{Lamp (shapenet)}: Lamps in greyscale, textures removed, grey background 
    \item \textbf{Telephone (shapenet)}: Telephones in greyscale, textures removed, grey background
    \item \textbf{Chair (shapenet)}: Chairs in greyscale, textures removed, grey background
    \item \textbf{Bench (shapenet)}: Benches in greyscale, textures removed, grey background
    \item \textbf{Sofa (shapenet)}: Sofas in greyscale, textures removed, grey background
    \item \textbf{Table (shapenet)}: Tables in greyscale, textures removed, grey background
    \item \textbf{Loudspeaker (shapenet)}: Loudspeakers in greyscale, textures removed, grey background
    \item \textbf{Display (shapenet)}: Display monitors in greyscale, textures removed, grey background
    \item \textbf{Cabinet (shapenet)}:  Cabinets in greyscale, textures removed, grey background
    \item \textbf{Car (shapenet)}: Cars in greyscale, textures removed, grey background
    \item \textbf{Watercraft (shapenet)}: Watercrafts in greyscale, textures removed, grey background
    \item \textbf{Airplane (shapenet)}: Airplanes in greyscale, textures removed, grey background
    \item \textbf{Abstract4 (shapegen)}: Procedurally generated non-semantic objects, most dissimilar match/non-matching object, greyscale, textures removed, grey background
    \item \textbf{Abstract3 (shapegen)}: Procedurally generated non-semantic objects, second-most dissimilar match/non-matching object, greyscale, textures removed, grey background
    \item \textbf{Abstract2 (shapegen)}: Procedurally generated non-semantic objects, intermediate dissimilarity between match/non-matching object, greyscale, textures removed, grey background
    \item \textbf{Abstract1 (shapegen)}: Procedurally generated non-semantic objects, second-most similar match/non-matching object, greyscale, textures removed, grey background
    \item \textbf{Abstract0 (shapegen)}: Procedurally generated non-semantic objects, most similar match/non-matching object, greyscale, textures removed, grey background
    \label{descriptions_of_categories}
\end{enumerate}

\begin{figure}[ht!]
\begin{center}
\includegraphics[width=.65
\textwidth]{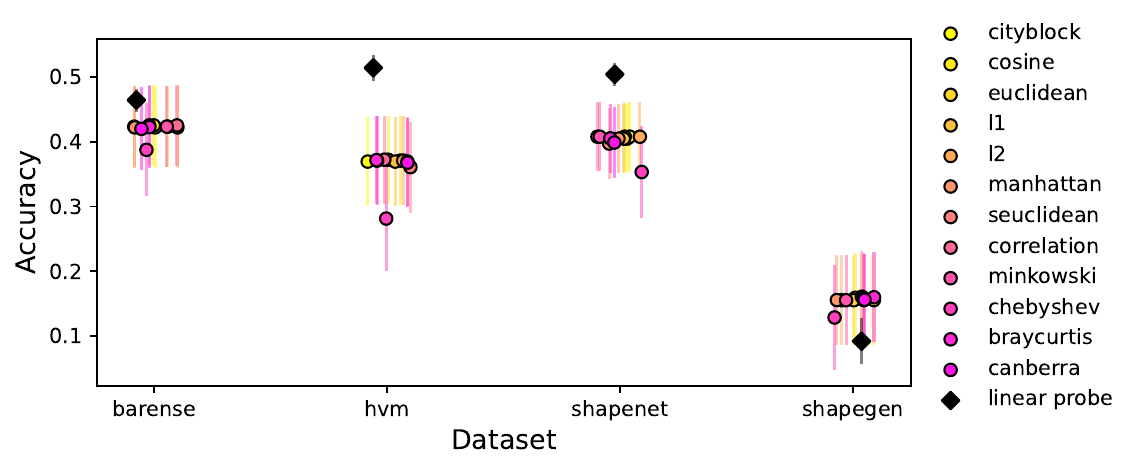}
\includegraphics[width=.3\textwidth]{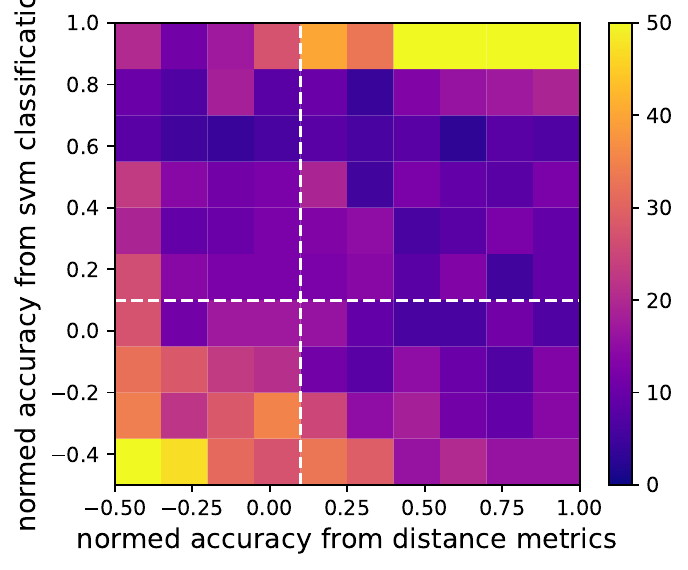}
\end{center}
\caption{\textbf{Comparing distance metrics to a lightweight linear readout.}}
\label{comparing_distance_metrics_scatter}
\end{figure}

\end{document}